\documentclass[manuscript]{IEEEtran}
\IEEEoverridecommandlockouts

\usepackage[T1]{fontenc}
\usepackage{amsmath,amssymb,amsfonts}
\usepackage{algorithmic}
\usepackage{graphicx}
\usepackage{textcomp}
\usepackage[dvipsnames,svgnames,table]{xcolor}
\usepackage{orcidlink}
\usepackage{hyperref}
\usepackage{wrapfig}
\usepackage{lipsum}
\usepackage{forest}
\usepackage{cleveref}
\usepackage{tabularx}
\usepackage{tabulary}
\usepackage{subcaption}
\usepackage{pdfcomment}
\usepackage{balance}
\usepackage{microtype}
\usepackage{orcidlink}
\usepackage[final,commandnameprefix=ifneeded,commentmarkup=uwave]{changes}

\crefformat{footnote}{#2\footnotemark[#1]#3}

\usepackage{tikz}
\usetikzlibrary{shapes.geometric,arrows,calc}

\definecolor{verylightgray}{gray}{0.90}

\tikzstyle{process} = [minimum width = 4.1cm, line width=1pt, minimum height = 0.6cm, rectangle, rounded corners, text centered, draw=black, fill=verylightgray, font=\scriptsize\sffamily]
\tikzstyle{arrow} = [thick,->,>=stealth, line width=1pt]

\usepackage{biblatex}
\def\BibTeX{{\rm B\kern-.05em{\sc i\kern-.025em b}\kern-.08em T\kern-.1667em\lower.7ex\hbox{E}\kern-.125emX}}

\begin{document}

\title{{\sc MetaMorph} -- A Metamodeling Approach for Robot Morphology
    \thanks{
        \added[id=rn]{
            This work was funded by the FET-Open Project \#951846 ``MUHAI – Meaning and Understanding for Human-centric AI'' by the EU Pathfinder and Horizon 2020 Program.
            This work has also been supported by the German Research Foundation DFG, as part of Collaborative Research Center 1320 Project-ID 329551904 ``EASE -- Everyday Activity Science and Engineering'', University of Bremen (\url{https://www.ease-crc.org}).
            The research was conducted in subproject ``P05-N -- Principles of Metareasoning for Everyday Activities''.
            %
            }}}


\author{\IEEEauthorblockN{Rachel Ringe}
\IEEEauthorblockA{\textit{Digital Media Lab} \\
\textit{University of Bremen}\\
Bremen, Germany \\
\href{mailto:rringe@uni-bremen.de}{rringe@uni-bremen.de}\orcidlink{0009-0005-4696-5873}}

\IEEEauthorblockN{Robin Nolte}
\IEEEauthorblockA{\textit{Digital Media Lab} \\
\textit{University of Bremen}\\
Bremen, Germany \\
\href{mailto:nolte@uni-bremen.de}{nolte@uni-bremen.de}\orcidlink{0009-0004-2975-6378}}

\IEEEauthorblockN{Nima Zargham}
\IEEEauthorblockA{\textit{Digital Media Lab} \\
\textit{University of Bremen}\\
Bremen, Germany \\
\href{mailto:zargham@uni-bremen.de}{zargham@uni-bremen.de}\orcidlink{0000-0003-4116-0601}}

\IEEEauthorblockN{Robert Porzel}
\IEEEauthorblockA{\textit{Digital Media Lab} \\
\textit{University of Bremen}\\
Bremen, Germany \\
\href{mailto:rringe@uni-bremen.de}{porzel@tzi.de}\orcidlink{0000-0002-7686-2921}}

\IEEEauthorblockN{Rainer Malaka}
\IEEEauthorblockA{\textit{Digital Media Lab} \\
\textit{University of Bremen}\\
Bremen, Germany \\
\href{mailto:rringe@uni-bremen.de}{malaka@tzi.de}\orcidlink{0000-0001-6463-4828}}
}

\maketitle
\IEEEpeerreviewmaketitle

\begin{abstract}

Robot appearance \replaced[id=rn]{crucially shapes}{plays a crucial role in shaping} \replaced[id=rn]{\emph{Human-Robot Interaction}}{human-robot interactions} (HRI)\replaced[id=rn]{ but is typically described via}{. Yet, to describe robot appearance, one typically relies on} broad categories like anthropomorphic, zoomorphic, or \replaced[id=rn, comment=this is the correct term from the paper; technomorphic was introduced by us I think?]{technical}{technomorphic}. More precise approaches focus almost exclusively on anthropomorphic features, which fail to \deleted[id=rn]{systematically} classify \deleted[id=rn]{the visual features of} robots across all types, limiting the ability to draw meaningful connections between \replaced[id=rn]{robot design}{a robot's design} and its effect on interaction.
\replaced[id=rn]{In response}{To address this gap}, we present \textsc{MetaMorph}, a comprehensive framework for classifying robot morphology. Using a metamodeling approach, \textsc{MetaMorph} was \replaced[id=rn]{synthesized from}{generated based on} 222 robots \replaced[id=rn]{in}{from} the IEEE Robots Guide, offering a structured method for comparing visual features. This model allows researchers to assess the visual distances between robot models and explore optimal design traits tailored to different tasks and contexts.
\end{abstract}


\section{Introduction}\label{sec:introduction}
\replaced[id=rn, comment=shortened; added a reference from the reviews]{Appearance critically shapes first impressions of robots \cite{Haring2016How}, and many studies highlight its role in \emph{Human-Robot Interaction} (HRI) \cite{riek2009how, strait2014let, natarajan2020effects, tenhundfeld2020robot}.}
{Appearance is one of the first factors humans notice when encountering robots \cite{Haring2016How}, setting the tone for interaction. Indeed, various studies have established that robot appearance affects \emph{Human-Robot Interaction} (HRI) \cite{riek2009how, strait2014let, natarajan2020effects}}
\replaced[id=rn, comment=shortened]{The importance of \linebreak appearance is widely acknowledged: HRI studies often group and compare robots via broad categories \cite{baraka2020extended} such as anthropo\-morphic, zoomorphic, or technical \cite{yanco2004classifying}. We argue that moving towards a systematic understanding of how specific design features influence HRI demands a more structured approach.}
{To understand these effects, other than acknowledging the importance of appearance, having a clear way of describing and classifying robot appearances could allow researchers to draw meaningful connections between a robot’s design and its impact on human interactions. However, robots used in HRI studies are often described using broad and unified categorizations \cite{baraka2020extended} such as anthropomorphic, zoomorphic, or technical, as suggested, e.g., by \mbox{\textcite{yanco2004classifying}}.}

\replaced[id=rn, comment=shortened; added more detail wrt to Phillips as promised in the rebuttal]{
Classifying robot appearance often relies on intuition rather than on a systematic analysis of visual features \cite{ABOT}. Regarding anthropomorphic robots, \textcite{ABOT} built a database of robots with at least one anthro\-pomorphic feature. In separate studies, participants then judged the robots' human-likeness and whether each robot exhibits $19$ specific features such as \emph{mouth}, \emph{gender}, and \emph{wheels}. The presence and absence of these features were shown to predict the averaged human-likeness score \cite{ABOT}, and to also outperform the human-likeness score in predicting certain effects, such as the type of job people envision assigning to a robot \cite{tenhundfeld2020robot}.
}
{What features exactly contribute to classifying a robot as anthropomorphic as opposed to the other groups is often based on intuition rather than on a systematic description of the robots' visual features and the relationships between them \cite{ABOT}. To address this problem, \mbox{\textcite{ABOT}} built a database of anthropomorphic robots and investigated what features influence the perception of human likeness. However, they limited their focus to anthropomorphic robots, as their database only consists of robots with at least one anthropomorphic feature.}

\replaced[id=rn, comment={shortened, as promised: a discussion of whether the whole equals the sum of its parts}]{
Moreover, \textcite{HWANG2013459} found that the shape of a robot's individual parts affects how humans perceive its personality. For example, a robot with a cylindrical head but human-like trunk and limbs was seen as the most extroverted (vs.\ robots with a similar silhouette but differently shaped parts). A simplistic classification of the overall robot cannot reflect this finding, i.e., that the whole does not equal the sum of its parts w.r.t.\ robot appearance.}
{Similarly, a classification of robot body types developed by \textcite{HWANG2013459} exclusively describes the different variations of anthropomorphic body shapes. In their approach, a robot shape consists of arms, legs, head, and trunk, each of which can be cylindrical, rectangular, or human-like. While their study evaluates the possible permutations generated by this description system, it does not account for robots that lack one or more of these components, such as those without arms or legs.}

\added[id=rn]{We give more examples in the \nameref{sec:Related_Work} but, in summary: details matter. Although existing research has made great steps, we still lack a  systematic model that covers the full range of robot appearances in detail. For instance, \textcite{ABOT} primarily focus on anthropomorphic features, and their feature selection method is not fully transparent. Similarly, a later study \cite{Löffler2020a}, focusing on zoomorphic robots, reused less than 40\% of \citeauthor{ABOT}'s features, to instead introduce animal-specific ones, but seemingly without a structured methodology. A more comprehensive model could reduce the risk of overlooking the effects of unconsidered robot types and features.
}

\deleted[id=rn, comment={I feel like this argument is essentially given already in the very first paragraph (the broad categories). To shorten the paper, I would completely cut it}]{
Currently, there is no systematic or universal method for describing a robot's visual features that applies across all categories without focusing primarily on anthropomorphic robots.
\mbox{\textcite{goetz2003matching}} found that user preferences for a humanoid versus a more machine-like, i.e., technomorphic, robot depend on the given task. Based on this,\mbox{ \textcite{richert2018anthropomorphism}} suggest that a technomorphic robot might be preferred in an industrial setting. \mbox{\textcite{wzietek2024imagination}} found that a more industrial or functional design might be preferred, even in service robotics. Additionally, various studies have explored the differences in interaction between anthropomorphic and non-anthropomorphic robots \cite{wzietek2024imagination, malle2016which, katz2014attitudes}. In such cases, only the anthropomorphic robot could be systematically described using existing models, while its counterpart could only be described by its absence of anthropomorphic features, which would lead to robots with distinctly different visual features, that could influence the interaction, sharing the same description. This highlights the need for a more comprehensive approach to describing robot appearances that captures the full range of visual features across diverse robot types.}
To address this gap, we introduce a novel framework to systematically describe a robot's visual features\added[id=rn]{: the \textsc{MetaMorph} model of robot morphology}.
Following a preliminary concept evaluation with a group of roboticists, we \replaced[id=rn]{synthesized \textsc{MetaMorph} by applying a metamodeling approach, considering $222$ robots from the \emph{IEEE Robots Guide} \cite{IEEErobotsguide} to account for a diverse range of existing robot types.}{developed the \textsc{MetaMorph} model of robot morphology using a metamodeling approach.}
\textsc{MetaMorph} concretizes the \emph{morphological features} of a robot, including its key physical parts (such as \deleted[id=rn]{arms, bodies,} grippers, tools, eyes, or wheels) that are visually distinguishable by humans.
\replaced[id=rn]{In that sense, \textsc{MetaMorph} may function as an inclusive, systematically compiled feature list to empower future studies.
Beyond that,}{Furthermore} the model describes the \emph{compositional relationship} of the morphological features, i.e., how they are connected or associated within the robot's structure to form a unified whole. 
\added[id=rn]{This allows for a novel comparison of appearances between robots via graph edit distance, based not only on presence and absence of features, but also on composition.}
\textsc{MetaMorph} was applied to the $222$ robots \added[id=rn]{from which it originated}, resulting in the creation of a comprehensive dataset detailing the visual features of these robots and the connecting relationships between them.
\added[id=rn]{This data could also improve accessibility, e.g., for the visually impaired, while supporting users, designers, and researchers.}


\section{Related Work}
\label{sec:Related_Work}
Various researchers have proposed taxonomies to categorize robot appearance.
In their review of social robots, \textcite{fong2003survey} introduce a taxonomy that describes robot morphology \replaced[id=rn]{as }{in terms of ``}anthropomorphic\deleted[id=rn]{''}, \deleted[id=rn]{``}zoomorphic\deleted[id=rn]{''}, \deleted[id=rn]{``}functional\deleted[id=rn]{''}, and \deleted[id=rn]{``}caricatured\deleted[id=rn]{''}. \replaced[id=rn]{\textcite{yanco2004classifying} proposed dropping the latter category and dividing robots classified as such across the other groups based on the specific traits their appearance was intended to exaggerate.}{\textcite{yanco2004classifying} adapted this taxonomy and revised the original classification by removing the category labeled ``caricatured.'' They proposed to integrate robots that previously fell under this category into the other existing categories based on the specific traits their appearance was intended to exaggerate.}
Since then, various frameworks \replaced[id=rn]{(e.g., \cite{gervasi2020conceptual,onnasch2021taxonomy, RANZ201899}) use either version to describe robot morphology.}{ have integrated sections to describe robot morphology or design, using either version of this taxonomy \cite{gervasi2020conceptual,onnasch2021taxonomy, RANZ201899}}.
Similarly, \textcite{shibata2004overview} suggests classifying robots \replaced[id=rn]{as}{by their resemblance to humans and animals, splitting them into} ``Human-Type'', ``Familiar Animal Type'', ``Unfamiliar Animal Type'', and ``New Character/Imaginary Animal Type''. \textcite{baraka2020extended} combine the approaches by \citeauthor{fong2003survey} and \citeauthor{shibata2004overview} \replaced[id=rn]{to split}{into a taxonomy that splits} robots into \replaced[id=rn]{``Bio-Inspired''}{Bio-Inspired Robots}, including various sub-categories of \replaced[id=rn]{``Human-Inspired'' and ``Animal-Inspired''}{Human- and Animal-Inspired robots}, as well as \replaced[id=rn]{``Functional'' and, to describe robots inspired by objects, apparatuses, and imaginary things, ``Artifact-Shaped''}{Functional Robots and Artifact-Shaped Robots that are inspired by objects, apparatuses, and imaginary things}.

While these works \replaced[id=rn]{focus}{have mainly focused} on broad categori\replaced[id=rn]{es}{zations}\deleted[id=rn]{ or groupings of robots}, others \replaced[id=rn]{go into more detail on}{have provided more detailed descriptions of} specific visual features across different robot\replaced[id=rn]{s}{ models}.
\added[id=rn]{We start with the ones briefly mentioned in the \nameref{sec:introduction}.
To analyze expectations of household robots, \textcite{ezer2008} analyzed study participants' drawings and descriptions of robots coded via 53 dimensions, including, among others, features specific to appearance.
\textcite[pp.~132--133]{ezer2008} explains that the coding scheme was constructed via a hybrid approach, top-down exploring ``characteristics of participants’ envisioned home-based robots'' and bottom-up ensuring ``that all commonalities in answers were accounted for'', but does not go into further detail.
Unfortunately, the methodology is therefore not replicable.
}

\added[id=rn, comment=I feel that this level of detail is necessary to distinguish us from previous work as promised in the rebuttal]{The feature list proved useful in a similar study by \textcite{phillips2017what}, who analyzed $155$ drawings of robots to explore peoples' expectations towards robot appearance.
They reused 16 features from \citeauthor{ezer2008}, and added 3 new ones (e.g., weapons) to ``better suit the drawing tasks in [their] studies'' \cite[p.~1216]{phillips2017what}.
Four final features account for machinery (e.g., wheels), one is an artifact (weapon), and the others appear to be associated with humans (coincidentally, also with animals, e.g., eyes).
\textcite{ABOT} later compiled the \emph{ABOT Database}, which includes $200$ existing robots with at least one human-like feature, applying their coding scheme \cite{phillips2017what} to each entry.
They showed that certain features reliably predict human-likeness as assessed by humans.
Thus, although its synthesis is not entirely transparent and based on imagined robots, the coding scheme still proved beneficial for studying real-world robotic designs in HRI.
A later study \cite{Löffler2020a}, however, indicates the limited extent to which this is possible, as the coding scheme required extensive revision to transfer \citeauthor{ABOT}'s study to zoomorphic robots.
In the end, only seven features were reused, and 20 were freshly introduced or translated from anthro-\linebreak pomorphic to zoomorphic (e.g., using claws instead of feet).
}

\deleted[id=rn]{\mbox{\textcite{phillips2017what}} analyzed $155$ drawings of robots collected across three studies to examine users' expectations of robot appearance. The authors drawings were coded using a scheme embodiment, human-likeness, assumed functionality, and interaction, as well as a distinction between human-like or mechanical appearance. While the coding scheme included more detailed descriptors for robot components, such as body parts and facial features, it was based on people's drawings rather than actual robot models. As a result, some features, like ``weapons,'' were included, which may be less relevant to actual robot models typically used in HRI studies.}

\deleted[id=rn]{\mbox{\textcite{ABOT}} later compiled the ABOT Database of $200$ robots with at least one human-like feature. They then conducted a study where participants were provided with $19$ humanoid features like hands or mouth and their definitions and then asked to decide which features were present in each robot. Afterward, they analyzed the data on feature presence to detect correlations and created a higher-level category system grouping features into surface look, body manipulators, facial features, and mechanical locomotion. This allows for a more detailed description of robot features. However, their chosen features are focused on human-like characteristics with very strict definitions.}
\textcite{Reeves2020Social} \replaced[id=rn, comment=shortened]{reviewed social robotics literature, compiling a dataset of $342$ robots, including images}{conducted an extensive literature review on robots discussed in social robotics research and collected a set of robots mentioned in these publications that were then analyzed in a follow-up study}. \replaced[id=rn]{They coded the robots by}{The robots were coded based on} \replaced[id=rn]{21}{various} attributes \added[id=rn]{in six categories (head, skin type and shape, communication ability, motion, gender, and age)}, some related to their appearance and others to their capabilities. These attributes were partly drawn from \replaced[id=rn, comment=shortened]{\citeauthor{ABOT}'s \cite{ABOT} features}{the features collected by \mbox{\textcite{ABOT}} for the ABOT Database} and partly \replaced[id=rn, comment=the previous statement is unprecise; they chose them before coding if I understand correctly]{chosen at the researchers' discretion}{selected by the researchers during the coding process}. \deleted[id=rn]{The collected features were then divided into six categories: head, skin type and shape, communication ability, motion, gender, and age.} \replaced[id=rn]{Most features provide broad rather than detailed descriptions and, apart from exceptions such as ``Animal Shape,'' are primarily tailored towards anthropomorphic robots.}{While the coding scheme included some options for describing non-anthropomorphic robots such as ``Animal Shape,'' most features were tailored to anthropomorphic robots and provided broad descriptors rather than detailed feature descriptions.} \deleted[id=rn]{The dataset compiled from their literature review contains images of the $342$ robots mentioned in their screened articles on social robotics, along with information on the robot features collected during their follow-up study.}

\added[id=rn]{\textcite{HWANG2013459}} \replaced[id=rn]{explored}{conducted a study to examine} how the \replaced[id=rn]{shape of individual parts}{shape of a robot} influences the emotions \replaced[id=rn]{invoked by}{experienced during interactions with} service robots. To achieve this, they \added[id=rn]{systematically} developed a \deleted[id=rn]{systematic}classification of ``overall robot shapes,'' drawing from \replaced[id=rn]{50}{a variety of} robots found in movies, cartoons, and internet searches. However, as most of these robots were humanoid or humanoid-adjacent, their classification was limited to \replaced[id=rn]{head, trunk, arms and legs}{humanoid body shapes}.

More recently, \textcite{seifi2023firsthand} compiled a dataset of 73 \deleted[id=rn]{different} robot hands\replaced[id=rn]{. In an online}{ and evaluated user perceptions of these hands in an online study using images. During the} study, users rated factors \replaced[id=rn]{such as}{like} perceived danger, gender, \replaced[id=rn]{and}{or} friendliness. In a follow-up \added[id=rn]{lab} study, they verified that the results \deleted[id=rn]{would} also apply to physical robots by evaluating a subset of 8 actual robot hands\deleted[id=rn]{ in a lab study}. The \deleted[id=rn]{collected} data was published in an online database \deleted[id=rn]{of robot hands} that includes design features\replaced[id=rn]{, e.g.,}{like} amount of fingers \replaced[id=rn]{and}{or} the presence of a thumb\added[id=rn]{,} as well as human ratings \replaced[id=rn]{of, e.g,}{such as} human-likeness \replaced[id=rn]{or}{and} creepiness.

\textcite{kalegina2018characterizing} \replaced[id=rn]{coded}{gathered} images and videos of $157$ robots featuring rendered faces \deleted[id=rn]{and systematically coded them} based on various facial characteristics. These included the presence of specific facial features, their colors, as well as the size, shape, and placement of each element. Additionally, the coding incorporated any physical features attached to the rendered faces, as well as a categorization of the robots' appearance into anthropomorphic, zoomorphic, and mechanical. This dataset was then used to conduct two studies—one investigating user preferences for different rendered faces and another exploring how facial features influence perceptions of both the face and the overall robot.

\replaced[id=rn]{A more technical approach for describing robots is}{Furthermore, robots can also be described using more standardized technical approaches, such as} the \emph{Unified Robot Description Format} (URDF)\replaced[id=rn]{,}{. URDF is} an XML-based file format developed as part of ROS (\emph{Robot Operating System}) to describe a robot's dynamics, kinematics, and geometry \cite{tola2024understanding}. It provides a framework for detailing various physical aspects of \replaced[id=rn]{robots}{a robot}, including the shape of joints\deleted[id=rn]{through a \texttt{<visual>} tag}. However, URDF's descriptive power is limited to basic geometric shapes \replaced[id=rn]{and}{or} 3D models. As a result, URDF provides only little abstraction from the physical form, which is why it is unsuitable for comparing conditions based on specific features in HRI studies. 


\replaced[id=rn]{In summary, the}{Overall, a review of} existing literature on robot appearance classification highlights \replaced[id=rn]{a}{the} need for more sophisticated models capable of comprehensively representing robot appearance. To address this gap, in this work, we introduce \textsc{MetaMorph}, a model of robot morphology developed using a metamodeling approach. \textsc{MetaMorph} offers a formal and systematic method for describing a robot's visual features, accommodating all robot types and providing a more inclusive and detailed framework for classification.

\section{Methodology}
\label{sec:Methodology}
To develop a comprehensive model \replaced[id=rn]{of}{to describe} robot morphology, we took a multi-step approach. \replaced[id=rn]{First, we conducted a focus group}{We first conducted an initial focus group} to receive feedback on \replaced[id=rn]{an initial}{the} model concept and what aspects \deleted[]{of robot morphology }to focus on\deleted[id=rn]{during data collection}. Next, we followed a workflow inspired by metamodeling to develop our model systematically.

\subsection{Focus-Group / Pre-Study}
As an initial step, we conducted a focus group with four roboticists to receive feedback on our approach, specifically what features should be collected in later coding steps and what other descriptors, e.g., the spatial location of the features the experts considered necessary. We chose roboticists as experts for this step due to their familiarity with various robot models. 
The focus group demographics can be found in \autoref{tab:DemographicsFocusGroup}.

\begin{table}[tb!]
\centering
\begin{tabulary}{\columnwidth}{|C|C|C|L|C|}
\hline
\textbf{ID} & \textbf{Gender} & \textbf{Age} & \textbf{Frequency of Working With Robots} & \textbf{Self-Rated Expertise (Scale of 1-7)} \\
\hline
P1 & Male & 40 & 2-3 times a week & 5 \\
\hline
P2 & Male & 32 & 2-3 times a week & 5 \\
\hline
P3 & Male & 30 & Everyday & 4 \\
\hline
P4 & Male & 30 & 2-3 times a week & 6 \\
\hline
\end{tabulary}
\caption{Demographics of the focus group participants}
\label{tab:DemographicsFocusGroup}
\end{table}

Before the focus group, inspirational material was created to spark and encourage discussion between the participants to gain better feedback on the necessary options and level of detail for the later more systematic and standardized concept collection during the metamodeling process. For this, two researchers screened more than 200 pictures of robots from the IEEE Robots Guide website to gain an overview of present visual features, which they collected as a list. 
In addition, they discussed various approaches to describe the spatial positions of these features on the different robots, such as zones, a graticule, or coordinates. In the end, we chose an approach that describes the location of different features using a four-zone system divided into ``Supra Zone'' (attached on top), ``Lateral Zone'' (attached on the middle/surrounding part), ``Infra Zone'' (attached on bottom), and ``Core'' containing the trunk of the robot after subtracting all attached appendages. 
This zone-based approach ensures that spatial descriptions are sufficiently flexible to accommodate a wide range of robot morphologies while maintaining a degree of specificity that enables meaningful comparisons. 
The previously collected features were grouped based on their typical zone location on the robots, e.g., arms and heads as Supra-Lateral Appendages and feet as Infra-Appendages.
As the initial task, experts were provided with more than 200 pictures of different robot models taken from the IEEE Robot Guide and asked to come up with a system to describe their visual features that could be applied to as many different robots as possible. They then compiled a list of categories and features deemed significant for differentiating robots, which can be summarized as follows: skin/texture, presence of a face and expressiveness, limb configuration (e.g., legs versus wheels, presence and number of arms and legs), mobility (stationary versus mobile), overall shape, and design style (biologically inspired versus industrial, anthropomorphic versus zoomorphic, cute versus uncanny). 

Next, experts were presented with the previously mentioned discussion materials - the list of collected and grouped features and an explanation of how to describe the spatial location of features using zones - and asked to provide feedback on this approach.
Participants agreed that developing a comprehensive model to describe robot morphology is highly valuable and showed interest in using it in their research after the process was completed. In the closing remarks, they praised the different detailed options to describe appendages, facial features, body shapes, and spatial descriptions of their locations. However, they disagreed on categorizing different features based on their typical location in the zones, citing the examples of possible robot designs, for example, made for aquatic or aerial environments where appendages like feet typically present in the infra zones might be attached to the supra zone and suggested categorizing appendages by type instead.

\subsection{Metamodeling}
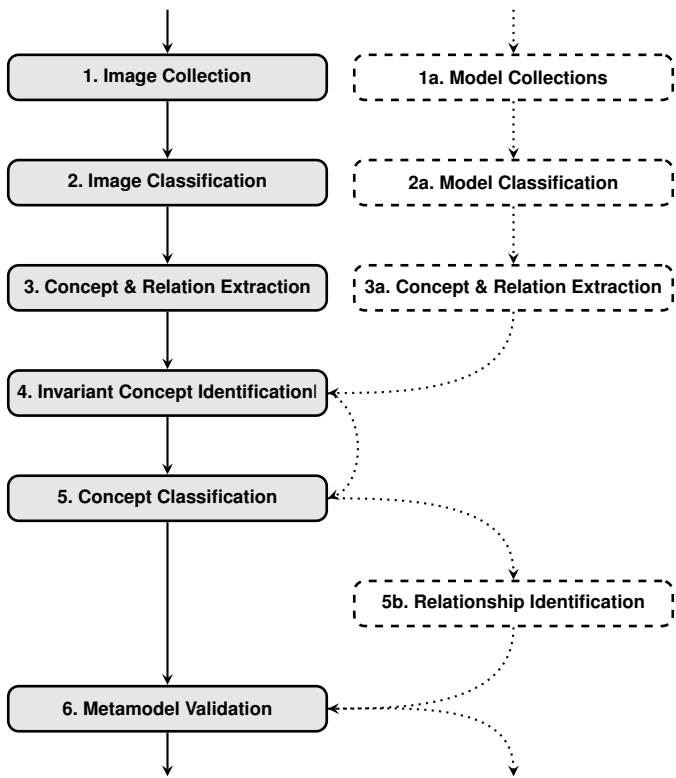
\begin{figure}[tb!]
    \centering
    \begin{tikzpicture}[node distance=1.4cm]
    \tikzstyle{arrow} = [thick, ->, >=stealth]
    \tikzstyle{darrow} = [thick, dotted, ->, >=stealth]
    \tikzstyle{dashedprocess} = [process, dashed, fill=white]
    \node (step1) [process] at (0,0) {\parbox{4cm}{\centering \textbf{1. Image Collection}}};
    \node (step2) [process, below of=step1] {\parbox{4cm}{\centering \textbf{2. Image Classification}}};
    \node (step3) [process, below of=step2] {\parbox{4cm}{\centering \textbf{3. Concept \& Relation Extraction}}};
    \node (step4) [process, below of=step3] {\parbox{4cm}{\centering \textbf{4. Invariant Concept Identification}l}};
    \node (step5) [process, below of=step4] {\parbox{4cm}{\centering\textbf{5. Concept Classification}}};
    \node (step7) [process, below of=step5, node distance=2.8cm] {\parbox{4cm}{ \centering \textbf{6. Metamodel Validation}}};
    
    \node (box1) [dashedprocess, right of=step1, xshift=3.2cm] {\parbox{4cm}{\centering \textbf{1a. Model Collections}}};
    \node (box2) [dashedprocess, below of=box1] {\parbox{4cm}{\centering \textbf{2a. Model Classification}}};
    \node (box3) [dashedprocess, below of=box2] {\parbox{4cm}{\centering \textbf{3a. Concept \& Relation Extraction}}};
    \node (step6) [dashedprocess, right of=step5, xshift=3.2cm, yshift=-1.4cm] {\parbox{4cm}{\centering\textbf{5b. Relationship Identification}}};
    
    \node (start) [coordinate, above of=step1, node distance=0.9cm] {};
    \node (start2) [coordinate, above of=box1, node distance=0.9cm] {};

    \draw [arrow] (start) -- (step1.north);
    \draw [darrow] (start2) -- (box1.north);

    \node (end1) [coordinate, below of=step7, node distance=0.9cm] {};
    \node (end2) [coordinate, right of=step7, xshift=3.2cm, yshift=-0.9cm] {};

    \draw [arrow] (step7.south) -- (end1);
    \draw [darrow] (step7.east) to[out=0, in=90] (end2);

    \draw [arrow] (step1) -- (step2);
    \draw [arrow] (step2) -- (step3);
    \draw [arrow] (step3) -- (step4);
    \draw [arrow] (step4) -- (step5);
    \draw [arrow] (step5) -- (step7);
    
    \draw [darrow] (box1) -- (box2);
    \draw [darrow] (box2) -- (box3);
    \draw [darrow] (box3) to[out=-90, in=0] (step4.east);
    \draw [darrow] (step4.east) to[out=-20, in=20] (step5.east);
    \draw [darrow] (step5.east) to[out=0, in=90] (step6.north);
    \draw [darrow] (step6) to[out=-90, in=0] (step7);
    
    \end{tikzpicture}
    \caption{The metamodeling steps. The dotted path follows the original approach used by \textcite{CaroPineres2014a}, while solid arrows represent our adaption for this paper.}
    \label{fig:metamodeling_steps}
\end{figure}

To design our model, we took inspiration from \emph{meta\-mod\-elling}, a \added[id=rn]{widely established technique} that synthesizes generalized models through qualitative analyses of sufficiently many exemplary models \cite{Sprinkle2010a}.
We roughly followed the seven steps \replaced[id=rn]{process}{of metamodel development} used by \textcite{CaroPineres2014a}, highlighted in \cref{fig:metamodeling_steps} as a dotted path.
\replaced[id=rn]{W}{However, w}e made two modifications for our purpose. First, as previously mentioned, we lack input models for robot morphology covering the full range of robot types, so we rely on image samples for steps 1 through 3 instead. Secondly, we refrained from identifying common relations (step 5b in \cref{fig:metamodeling_steps}) as we were not trying to paint a representative picture of how concepts \emph{typically} interact as is the case, e.g., with software metamodels.
Since we aim for a description scheme to be instantiated individually for each robot, which may have any configuration imaginable, it makes little sense to restrict the scheme to typical relations such as that hands must be connected to an arm.
The solid path in \cref{fig:metamodeling_steps} shows an overview of the final process applied here. We describe the individual steps in greater detail below.

\subsection*{Step 1 -- Image Collection}
\label{sec:step1}
This step involves preparing a suitable set of robot images for subsequent use. The set should represent a diverse range of robots and consist of full-body photos, each depicting a single robot, preferably in frontal to three-quarter view.
Most datasets mentioned in \Cref{sec:Related_Work} proved unfit for our cause as they only include robots that are at least partially anthropomorphic (ABOT Database \cite{phillips2017what}), focus on specific robot body parts (Dataset of Rendered Robot Faces \cite{kalegina2018characterizing}; Robot Hand Database \cite{seifi2023firsthand}), or are not available anymore via the links provided in the associated publications \added[id=rn]{and the authenticity of other versions cannot be verified}  (Dataset of Rendered Robot Faces \cite{kalegina2018characterizing}; Stanford Social Robot Dataset \cite{Reeves2020Social}).

Instead, we used the IEEE Robots Guide \cite{IEEErobotsguide}\replaced[id=rn]{, which}{. This dataset was ideal as it} includes various robot types\deleted[id=rn]{ and is regularly updated with new entries}. Although first published in 2018, the guide is continuously maintained, offering comprehensive coverage of major robot projects through high-quality images and videos. At the time of data collection (May 2024), the website featured 259 heterogeneous robots from various countries and companies created between 1961 and 2024.

A systematic exclusion process to ensure the dataset's relevance, then applied the following four \textbf{exclusion criteria}:
\begin{itemize}
    \item
    \textit{Without full-body images} (14 robots removed):
    Analyzing the complete appearance is impossible without full-body imagery, compromising our classification framework.
    \item 
    \textit{Cars and planes} (11 robots removed):
    We excluded robots classified as cars or planes in the database because their appearance is largely dictated by their vehicular function rather than their autonomous robotic characteristics.
    Robotic cars and airplanes are fundamentally designed for transportation rather than direct human interaction or independent, adaptive behavior in dynamic environments.
    Their primary design constraints focus on aerodynamics, mechanics, and functional efficiency related to transportation, which are different from those of robots designed for HRI, service, or industrial tasks. 
    \item
    \textit{Exoskeletons} (6 robots removed): We excluded exoskeletons as they are augmentative devices rather than standalone robots.
    While robotic, these devices lack other robot types' independent visual identity. 
    \item
    \textit{Modular robot parts} (8 robots removed): We excluded modular robot parts as they \deleted[id=rn]{do not represent complete robots with cohesive appearances.
    They} are \replaced[id=rn]{meant}{designed} to be assembled into various configurations but lack a singular, unified form \added[id=rn]{with cohesive appearances}.
\end{itemize}

After applying the exclusion criteria, 220 robots remained. However, two robots, Aquanaut \cite{Manley2018a} and HRP-2 \cite{Isozumi2004}, have the ability to \emph{transform}, i.e., they have two visually distinct forms in different states. As these represent important variations in appearance, we included both additional forms individually. The final dataset comprised 222 robots.

\subsection*{Step 2 -- Image Classification}
We randomly partition the prepared image set into \replaced[id=rn]{\emph{template samples}}{a \emph{template set}} (TS) and \replaced[id=rn]{\emph{validation samples}}{a \emph{validation set}} (VS)\replaced[id=rn]{, 111 images each}{ of nearly equal size (TS: 111; VS: 111)}. The TS \replaced[id=rn]{are}{is} reserved for compiling the metamodel in Steps 3\replaced[id=rn]{--}{ through }6 and the VS for validating the model in Step 7. The supplementary material\footnote{\label{github}All supplementary materials, including the {\sc Meta\-Morph} taxonomy, data gathered during the metamodeling process, and the dataset of annotated robots, are publicly available at \url{https://github.com/RRachelRR/MetaMorph}.} lists which set includes which robots.

\subsection*{Step 3 -- Concept and Relation Extraction}
\begin{figure}[tb]
     \centering
     \includegraphics[width=0.8\linewidth]{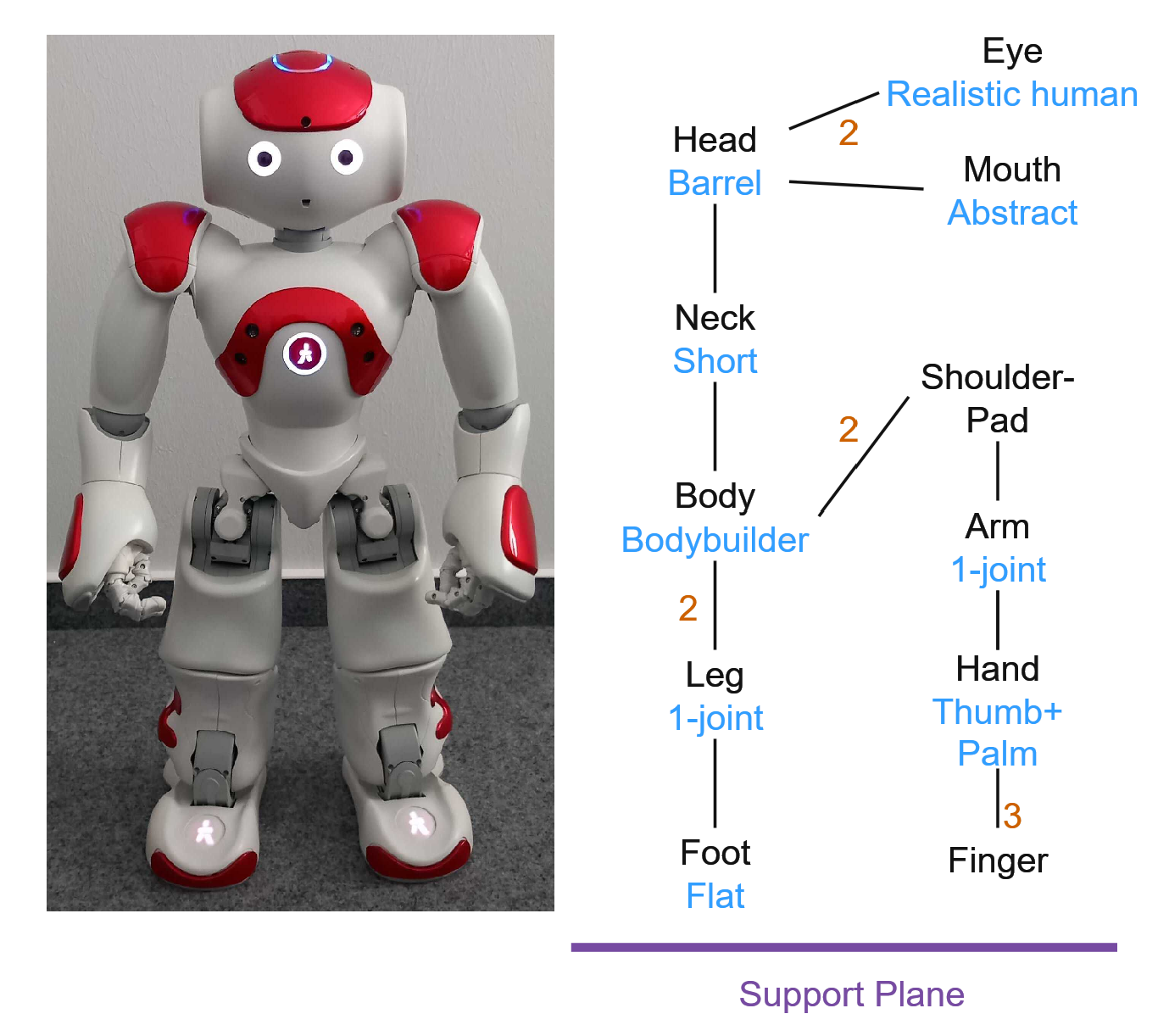}
     \caption{Picture of the NAO \cite{Gouaillier2008a} robot and the corresponding labeled graph as extracted in step 3. Black text represents a morphological subdivision, blue text represents morphological descriptors, and orange numbers summarize multiplicities of branches (assuming the \textit{Body} being the graph's root).}
     \label{fig:step3graph}
\end{figure}

\replaced[id=rn]{T}{For Step 3, t}wo researchers jointly analyzed \added[id=rn]{the TS, manually translating} each \deleted[id=rn]{robot} image\deleted[id=rn]{ from the TS to manually translate it} into an undirected graph with labeled vertices. Each vertex represents exactly one morphological subdivision and, possibly, any number of additional morphological features that provide further \deleted[id=rn]{visually} relevant information, \replaced[id=rn]{e.g., }{such as }shape \replaced[id=rn]{and}{or} level of detail. Together, these form the set of extracted concepts. Edges represent the structural connection between features (i.e., the extracted relations). Fig.\ \ref{fig:step3graph} shows an example of such a graph. When the experts disagreed, they produced separate graphs that a third independent HRI expert judged to reach a consensus. 

The extracted morphological features are based on what the researchers identified as apparent instead of existing feature classifications for different reasons:
The feature list by \textcite{kalegina2018characterizing} is unavailable.
\citeauthor{Reeves2020Social}'s features \cite{Reeves2020Social} stay fairly general such that, for example, mouth, eyes, and nose are abstracted away to ``has face.''
\deleted[id=rn]{\textcite{phillips2017what} extract more specific features; however, these are based on people's drawings instead of actual robots.}
The \replaced[id=rn]{19}{extensive list of} anthropomorphic features used \replaced[id=rn]{by}{in} \textcite{phillips2017what,ABOT} \replaced[id=rn]{are}{is} primarily associated with somewhat restrictive definitions. For example, viewing skin as ``a thin layer of tissue covering almost the entire body'' \cite{ABOT} excludes robots that feature little skin, e.g., when only the face is covered.
Similarly, their definition of a mouth as ``a large opening located on the lower [\dots] face'' \cite{ABOT} does not apply to mouths that are painted on, rendered, or represented by other physical components.

We consider \citeauthor{fong2003survey}'s \cite{fong2003survey} classification into anthropomorphic, zoomorphic, and functional to be too general as descriptors for the entire robot but use them as additional descriptors for morphological features.

Inspired by feedback from the focus group, the researchers \deleted[id=rn]{also }associated each robot with a covering and an overall silhouette. We decided to adapt \citeauthor{Reeves2020Social}'s \cite{Reeves2020Social} approach to describe coverings\replaced[id=rn]{, coding both}{. They describe coverings by a combination of} their material (e.g., plastic, metal, fur) and the visibility of mechanics.

        %
        %
        %


\subsection*{Step 4 -- Invariant Concept Identification}
Next, we sort the morphological features and annotations extracted in Step 3 in descending order by the number of robot graphs in which they occur. We extracted $133$ features, each occurring in an average of $\approx 7.6$ robots with a standard deviation of $\approx 14.8$.
Seven features occurred particularly often, measured by a z-score $\geq 2$: \textit{Body} (91 robots), \textit{Head} (74), \textit{Arm} (60), \textit{Leg} (44), \textit{Neck} (42), \textit{Eye} (39), and \textit{Camera} (38)\footnote{\added[id=rn]{Based on the features presented here, it could be assumed that the TS consists mostly of humanoid robots. However, others such as \href{https://robotsguide.com/robots/chico}{\emph{Chico}} \cite{Guizzo2008a}, which has an arm, a head, and legs but is more of an excavator shape, also contribute.}}.
We cannot make any statistically relevant observations regarding rare features since, with the given distribution, the 69 features and 34 annotations that occur only once have a z-score of $\approx -0.45$.
Following the metamodeling approach, the latter still qualify as candidates for merging with more general or similar but more common concepts in Step 5\footnote{Since our goal is to develop a complete model, we slightly deviate from \citeauthor{CaroPineres2014a}'s \cite{CaroPineres2014a} approach, which excludes rare concepts entirely.}. We also bookmark them for careful validation in Step 7.

\subsection*{Step 5 -- Concept Classification}
In this step, we create an ontological taxonomy of the morphological features, annotations, and silhouettes. We identified equivalency, subsumption, and sibling relationships between morphological features by comparing freshly produced definitions for each feature. These definitions depend on the visual appearance rather than physical form, e.g., a \textit{Mouth} ``visually resembles a human or animal mouth to the extent that it is readily perceived as such by human observers [\dots].''

Equivalent morphological features were unified.
For sibling classes lacking a parent that adequately differentiates them from other morphological features, we introduce new superclasses inspired by anatomical categories from the Uberon ontology \cite{Mungall2012Uberon} of multi-species anatomy. For instance, \textit{Head}, \textit{Limb Segment}, and \textit{Neck} were grouped under the novel parent class \textit{Connecting Subdivision}, reflecting their shared affordance of linking the robot's core to other morphological subdivisions.
\deleted[id=rn]{Note that some morphological subdivisions were renamed in this process; for example, an ``appendage'' became a \textit{Connecting Subdivision}.}
\added[id=rn]{The terminology is chosen to be unambiguous, e.g., to specify that we are referring to the part between shoulder and hand, rather than an entire arm.
This adds a layer of abstraction to manage the extensive feature list but does not take away from them. }
The resulting taxonomy was then refined by reviewing each candidate for potential merging (as identified in Step 4) and dropping those with a sufficiently specific parent class, e.g., all subclasses of \textit{Tool}.

Additionally, to keep the model consistent, some features initially listed as subdivisions have been reclassified as descriptors, namely finger and foot configurations such as the number of fingers or special shapes (e.g., \textit{Mitten} and \textit{Paw}).
This is because we also consider shapes and joint amounts as morphological descriptors rather than subdivisions.
Last, we dropped subdivisions that we retrospectively identified as not part of the robot, e.g., soil used by a gardening robot.


\subsection*{Step 6 -- Metamodel Validation}
\added[id=rn, comment=maybe move this to the discussion/limitations?]{Metamodeling prioritizes validation through experts instead of by users. While previous studies (e.g., \cite{HWANG2013459,phillips2017what,Löffler2020a,tenhundfeld2020robot}) analyzed feature influence on perception through participant feedback, this is distinct from directly validating a classification system. The validation that we present in this paper represents the first systematic validation of a model specifically for robot appearance (of course, further user-focused validation is planned to deepen evaluation and refine applicability).
Our methodology is as follows:
}

\replaced[id=rn]{The same researchers as in step 3}{Two researchers} repeated the process of translating robot images into labeled graphs, but this time using the \deleted[id=rn]{images from the }VS, guided by the taxonomy of morphological features that was derived from the TS in steps 3 through 5. The researchers were tasked to note any shortcomings for later revision of the taxonomy.
In most cases, the morphological features of the VS robots could be sufficiently described with the compiled concepts. In the following, we briefly explain the issues identified in the validation process and our repairs:

\begin{itemize}
    \item A number of morphological subdivisions that were merged with Tool due to only occurring once, such as Lamps and Syringes, occurred multiple times in the VS. These were then reintroduced into the taxonomy.
    \item The VS contained an accumulation of specific morphological subdivisions, such as suction cups, pulley wheels and prominent cable bundles, which did not occur at all in the TS. Corresponding concepts were integrated.
    \item Certain shapes and silhouettes could not be adequately described. Accordingly, we added missing concepts such as hemisphere and Insect-Base-Hybrid.
\end{itemize}

We decided not to test the existing concepts for relevance to the VS, as any exclusion thereof would have directly led to robots that could not be fully described by the model. The resulting model is detailed in the following section.

The final taxonomy is provided\cref{github} as an OWL \cite{Hitzler2012a} ontology (that is currently not aligned to any foundational model).

\section{The \textsc{MetaMorph} Model}

\begin{figure}[htbp!]
    \centering
    \includegraphics[width=0.5\textwidth]{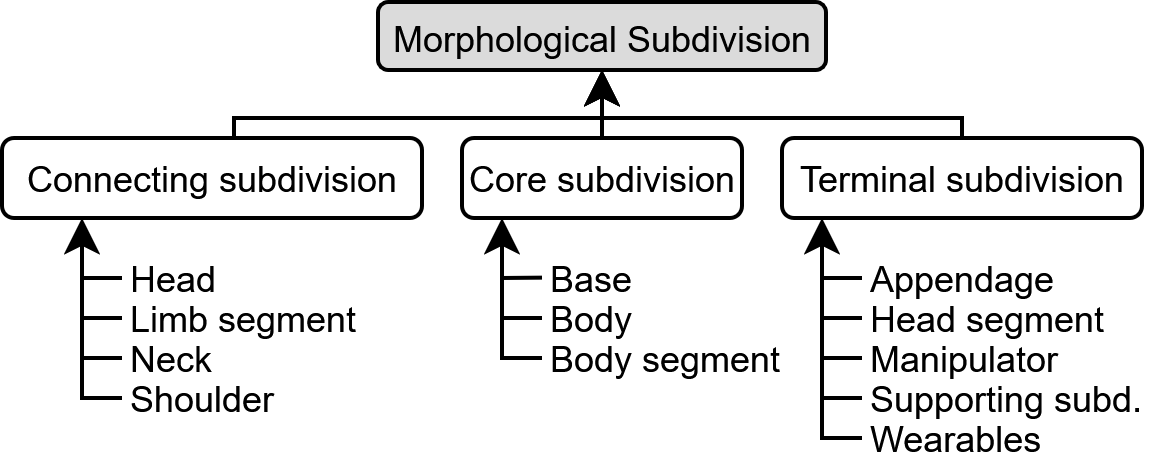}
    \caption{Part of the constructed taxonomy's branch of \emph{Morphological Subdivisions}. Due to limited space, concrete \emph{Terminal subdivisions}, such as \emph{Tail} or \emph{Tool} are omitted, and can be found in the supplementary material\cref{github}.}
    \label{fig:subdivision_taxonomy}
\end{figure}

\begin{figure}[htbp!]
    \centering
    \includegraphics[width=0.5\textwidth]{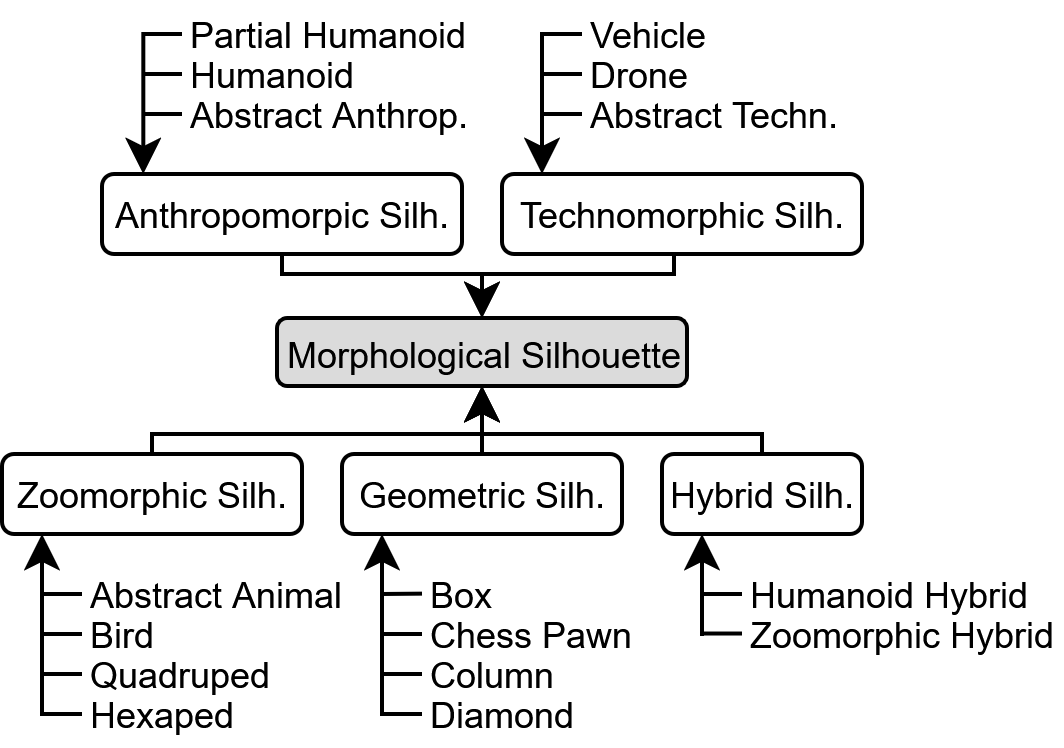}
    \caption{Part of the constructed taxonomy's branch of \emph{Morphological Silhouettes}. Due to limited space, some concepts are omitted, but can be found in the supplementary material.}
    \label{fig:subdivision_taxonomy2}
\end{figure}

The final version of the \textsc{MetaMorph} model provides a detailed way to describe robot morphology. It is split into morphological subdivisions, descriptors that can be applied to the features, and morphological features that describe the whole robot, such as coverings and silhouettes. We define three types of morphological subdivisions (visualized in Figure \ref{fig:subdivision_taxonomy}):

\begin{itemize}
    \item \textbf{Connecting subdivision:} A morphological subdivision of the robot that connects two or more other morphological subdivisions, e.g., a \textit{Neck} may connect \textit{Torso} and \textit{Head}. \textit{Connecting Subdivisions} are further divided into \textit{Head}, \textit{Neck}, \textit{Shoulder}, and \textit{Limbs}.
    \item \textbf{Terminal Subdivision:} A morphological subdivision that connects to only a single other morphological subdivision such as \textit{Hands}, \textit{Tails}, \textit{Eyes}, and \textit{Wheels}. These subdivisions are split into five groups. \textit{Head segments} describe various facial features by the region of the face in which they are placed (high, mid, and low), as well as screens, hair, and sensor arrays. \textit{Hands}, \textit{Grippers}, \textit{Tools}, \textit{Knobs}, and \textit{Suction Cups} form the \textit{Manipulator} group. \textit{Wearables} include \textit{Clothing} and \textit{Accessories}, while the \textit{Appendages} category includes miscellaneous options such as \textit{Antenna}, \textit{Backpack}, \textit{Handle}, and \textit{Wing}, among others. Lastly, the \textit{Supporting Subdivisions} describe various categories of components that connect the robot to the support plane. This category offers a variety of options used in air, ground, and water environments, as well as rope connectors.
    \item \textbf{Core Subdivision:} A morphological subdivision that is at the core of the robot, appearing to give basic structural stability and directly connected with \textit{Connecting Subdivisions} (such as \textit{Arms} and \textit{Legs}) and \textit{Terminal Subdivisions}, typically \textit{Appendages} such as \textit{Tails} or \textit{Fins}. This category includes \textit{Bases}, \textit{Bodies} (including its more specific subcategory, \textit{Torso}), and \textit{Body Segments} such as \textit{Thorax} and \textit{Abdomen}.
\end{itemize}

In addition to the morphological subdivision, the model also contains \textit{Morphological Descriptors} that can be applied to a subdivision to provide additional details about its visual appearance. \textit{General Descriptors} can apply to any subdivision, while \textit{Subdivision-Specific Descriptors} only apply to certain kinds of subdivision, e.g., \textit{Hand or Gripper configuration} can only be used to describe the appearance of a Hand-subdivision. The first subcategory of \textit{Morphological descriptors} addresses \textit{Morphism} - the likeness of a subdivision to either its human, animal, or technical equivalent. Furthermore, the morphological descriptors allow for descriptions of the \textit{Degree of realism} with which a certain subdivision, e.g., an eye, is represented. This can be described in terms of hyperrealistic (an extremely lifelike depiction, e.g., an accurate copy of a human eye as seen on androids like the Geminoid DK), realistic (a detailed depiction that mimics the real object but slightly simplified, e.g., an almond-shaped eye with pupil and eyelids), abstract (a very simplified depiction, e.g., a dot representing an eye or a rendered cartoon eye) and symbolic (a technical component of a similar shape representing a feature, e.g., a round camera lens representing an eye). Additionally, the \textsc{MetaMorph} model provides a descriptor for a subdivision's shape in terms of a variety of geometric forms identified from the TS and VS.

Coverings are described based on the level of coverage they provide for the robot's mechanics and their materials. They are split into \textit{mechanics fully covered}, \textit{mechanics partially covered}, and \textit{mechanics fully visible}, and then they are further divided by the different combinations of covering materials identified during the metamodeling process.

The \textsc{MetaMorph} model describes the overall shapes of robots as \textit{Morphological Silhouettes} visualized in Fig.  \ref{fig:subdivision_taxonomy2}. They are subcategorized into \textit{Anthropomorphic} (resembling humans), \textit{Zoomorphic} (resembling animals), and \textit{Technomorphic} (resembling technological devices or vehicles), as well as \textit{Geometric} (best described by a geometric shape) and \textit{Hybrid Silhouettes} that combine two of the other categories.

\begin{figure*}[htbp!]
    \centering
    \begin{subfigure}[b]{0.32\textwidth}
        \centering
        \includegraphics[width=\textwidth]{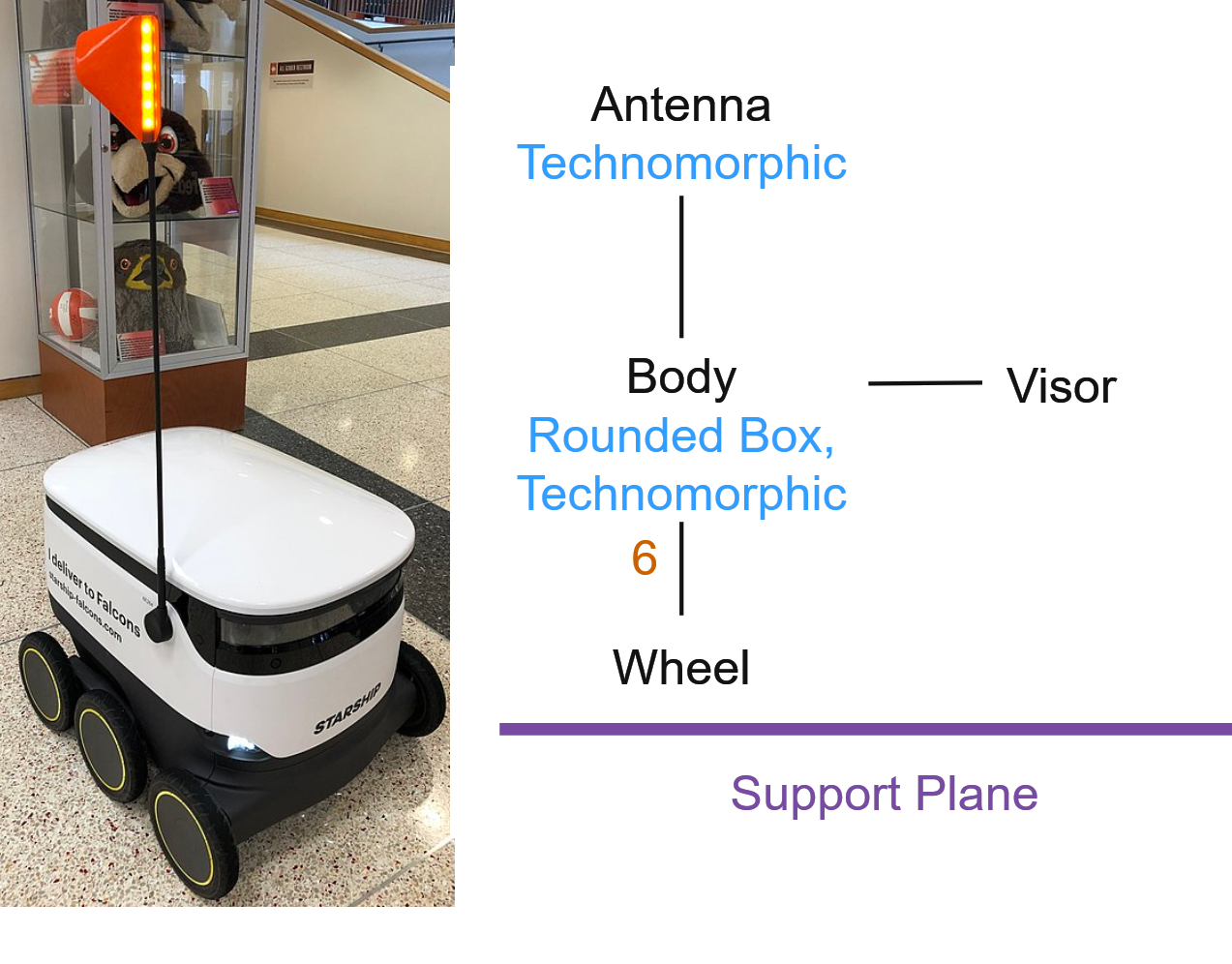}
        \caption{The robot \emph{Starship} (Image adapted from \cite{StarshipImage}).}
        \label{fig:starship}
    \end{subfigure}
    \hfill
    \begin{subfigure}[b]{0.32\textwidth}
        \centering
        \includegraphics[width=\textwidth]{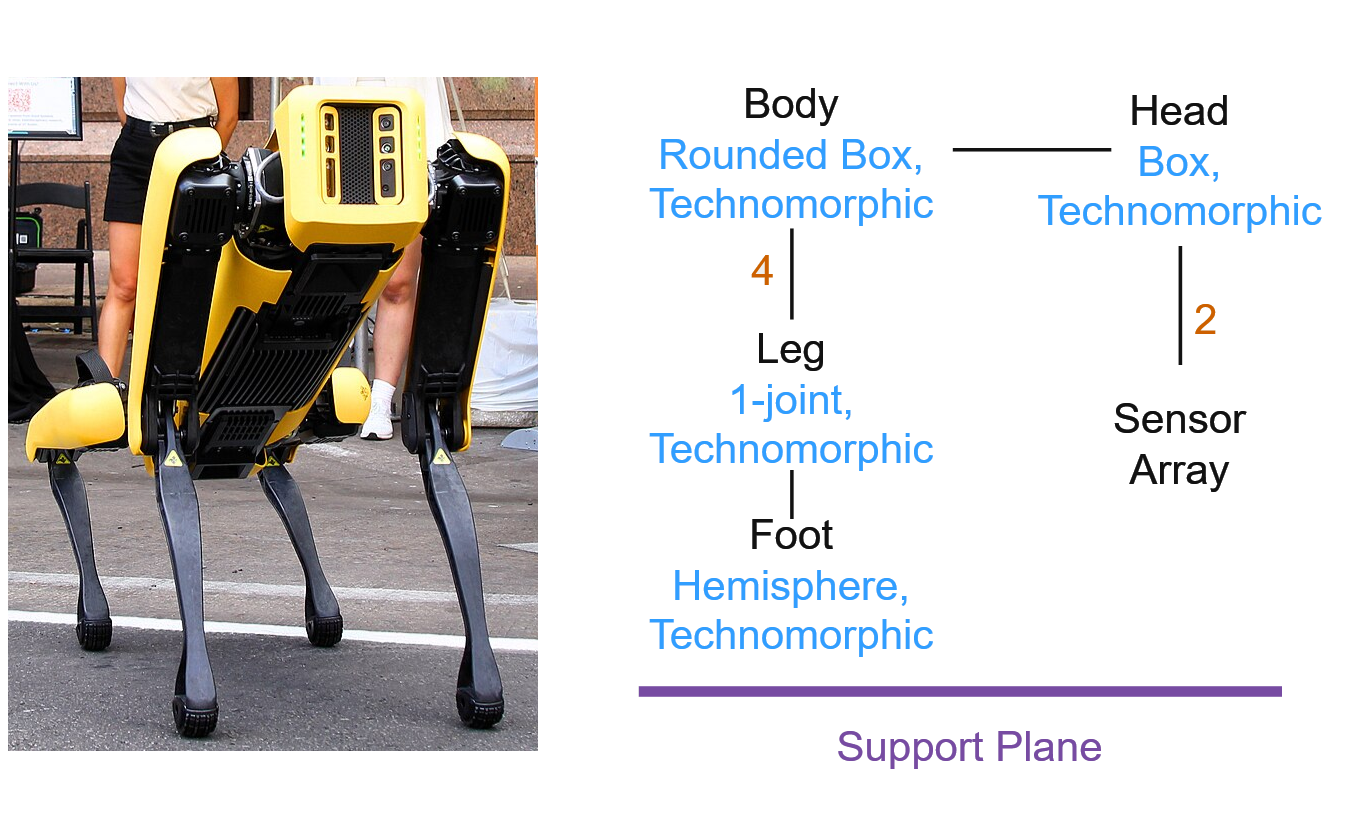}
        \caption{The robot \emph{Spot} (Image adapted from \cite{SpotImage})}.
        \label{fig:spot}
    \end{subfigure}
    \hfill
    \begin{subfigure}[b]{0.32\textwidth}
        \centering
        \includegraphics[width=\textwidth]{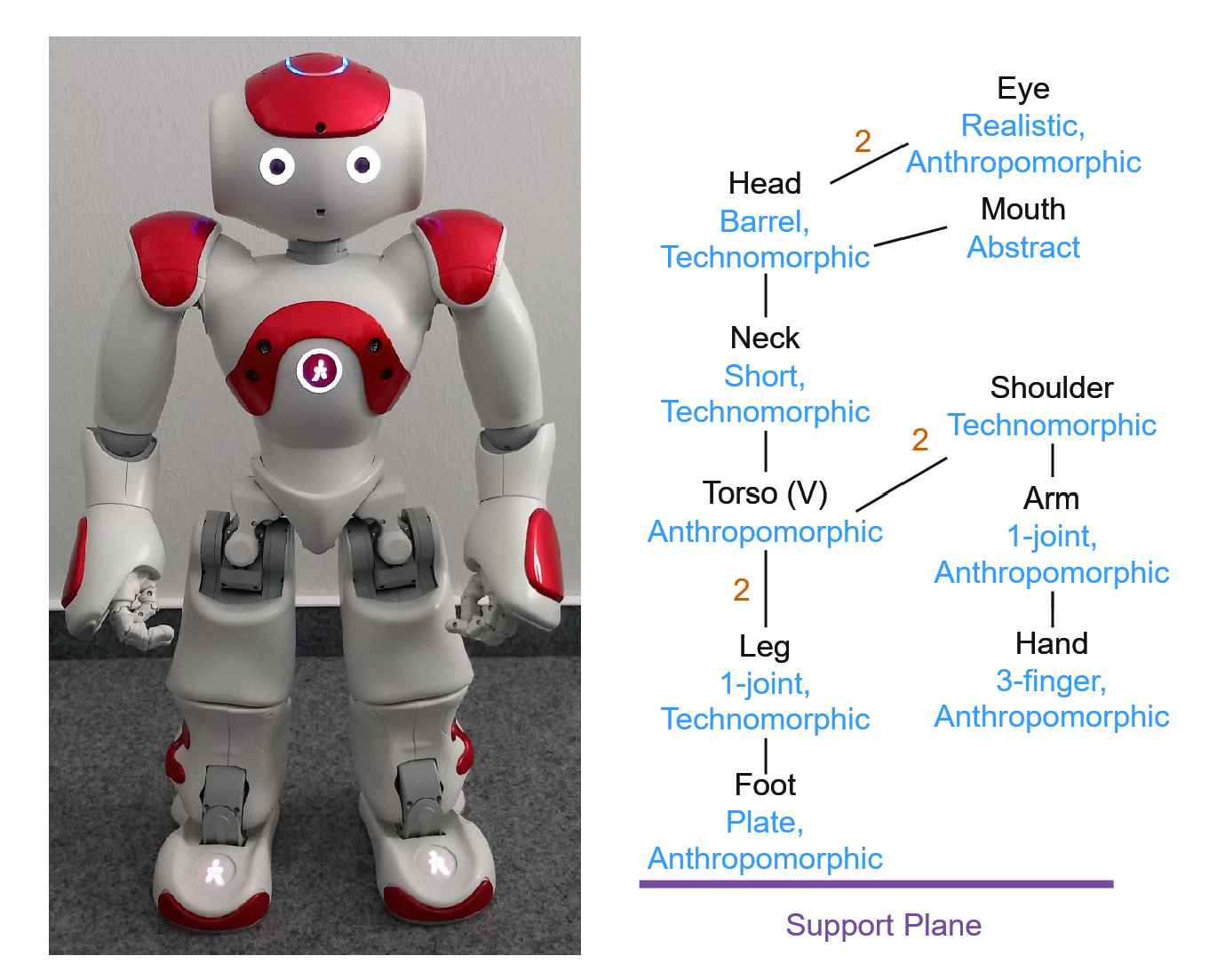}
        \caption{The robot \emph{Nao} \cite{Gouaillier2008a}}
        \label{fig:nao}
    \end{subfigure}
    \caption{Different robots (each left) and their associated \textsc{MetaMorph} description graphs (each right).}
    \label{fig:finalGraphs}
\end{figure*}

Using the \textsc{MetaMorph} model, a robot can be described either by compiling a list of present visual features, including descriptors, or by creating an undirected, labeled graph that contains the morphological features as nodes, their descriptors as additional labels, and the edges to describe the connections between parts. Figure \ref{fig:finalGraphs} shows final descriptions for three different robots: the NAO \cite{Gouaillier2008a} robot, for which also the earlier version of the graph was shown in Figure \ref{fig:step3graph}, the Starship robot, and the Spot robot.

\subsection{Dataset and Distance Calculation}

\replaced[id=rn]{We collected data on all robots from the TS and VS by applying the final {\sc MetaMorph} metamodel. For each robot, t}{Based on the validated metamodel, a dataset with descriptions for all the robots from the Validation and Test set was collected and is available on GitHub\cref{github}. T}he dataset\cref{github} contains a list of \replaced[id=rn]{features present}{visual features for every robot}, \replaced[id=rn]{plus}{including} more detailed descriptors and the connecting relationship\deleted[id=rn]{ tuple}s to construct the graph describing the robot's morphology. The \replaced[id=rn]{data enables visual distance metrics}{information in this dataset allows the calculation of visual distances} between \deleted[id=rn]{different} robot\deleted[id=rn]{model}s, either \replaced[id=rn]{via the Jaccard index, simply}{using a boolean array method} comparing \replaced[id=rn]{features,}{the presence of features} or \added[id=rn]{via} a graph edit distance approach, which \replaced[id=rn]{accounts for structural composition}{also includes the relationships between features and multiplicities}.
\replaced[id=rn]{\Cref{table:distance} contains the results for the robots shown in fig.\ \ref{fig:finalGraphs}.}{The visual distance between the robots shown in Figure \ref{fig:finalGraphs} was calculated using the Python package NetworkX\footnote{\url{https://networkx.org/}} to calculate the graph edit distance as well as the package scikit-learn\footnote{\url{https://scikit-learn.org/}} to calculate the Jaccard score between lists of features. Results can be seen in Table \ref{table:distance}.}

\begin{table}[tb!]
    \centering
    \begin{tabularx}{\linewidth}{|l|X|X|X|}
        \hline
        \textbf{Algorithm} &  \textbf{Starship/Spot} & \textbf{Starship/Nao} &  \textbf{Spot/Nao}\\ 
        \hline
        Jaccard index & 0.125 & 0.0 & 0.07\\ 
        \hline
        Graph edit distance & 20 & 29 & 20\\
        \hline  
    \end{tabularx}
    \caption{Distances between the {\sc MetaMorph} description graphs from \cref{fig:finalGraphs}.
    The Jaccard index was calculated via \emph{scikit-learn} \cite{scikit-learn}, and the graph edit distance using \emph{NetworkX} \cite{Hagberg2008a}.
    }
    \label{table:distance}
\end{table}

These calculations serve purely as a proof of concept for the use case of distance calculation between the appearances of two robots since the standard implementations were used where every difference between the feature lists and graph editing operation is treated equally. This means the difference between two eye subdivisions with different descriptors would be the same as the difference between an eye and a leg subdivision. Further research is needed to determine the optimal weights of the changes between different subdivisions and descriptors, as well as the addition and deletion of subdivisions to allow for an accurate calculation of visual distance.

\section{Discussion}
This work proposes a comprehensive framework for classifying robot morphology. Using a metamodeling approach, we developed the \textsc{MetaMorph} model, which offers a structured method for classifying and comparing robots' visual features.

Existing models \replaced[id=rn]{of}{to describe} robot appearance often \replaced[id=rn]{group}{provide broad descriptions by grouping} robots into a few \added[id=rn]{broad} categories. 
While some approaches \replaced[id=rn]{go into more detail}{offer more detailed descriptions}, these tend to focus primarily on anthropomorphic features or \deleted[id=rn]{provide }specific \deleted[id=rn]{descriptors for }components such as \replaced[id=rn]{faces and hands}{facial characteristics}. With the \textsc{MetaMorph} model, we aimed to build on this foundation by incorporating non-anthropomorphic features and offering a more granular level of detail across various robot types.
In developing our taxonomy, we carefully assessed the integration of existing models, leading to the inclusion of the anthropomorphic-zoomorphic-technical taxonomy \cite{yanco2004classifying} and a description of coverings inspired by \textcite{Reeves2020Social}. Even though we \replaced[id=rn]{constructed}{elected to construct} our own taxonomy based on the metamodeling process for the remaining descriptions, other similarities to existing work can be found in parts of the model -- especially the categorization of basic facial features. 


\replaced[id=rn]{
Note that even though its feature list is extensive, \textsc{MetaMorph} may occasionally generate identical descriptions for two different robots. For example, this first version describes a robot with only a hand covered with skin as having the same morphological covering as one with only a face covered. This may be problematic for appearance studies where such robots are used for different conditions. A future iteration could deviate from \textcite{Reeves2020Social} by assigning coverings to individual parts rather than the entire robot, as is already the case with other descriptors. However, it is unlikely that such problems can be completely avoided with a generally applicable model, since edge cases can require arbitrarily detailed descriptions to solve, e.g., of color, size or eye distance.}{
It should be recognized that, while we do provide a detailed list of visual features, including descriptors for every robot, there will still be cases where two robots would receive the same description while being used in different conditions in a study exploring appearance. For example, in the current version of the \textsc{MetaMorph} model, a robot whose hands are covered by skin while the rest is covered in chassis would have an identical description to the same model with only the face covered in skin.While adjustments could be made to future versions of the model to add additional descriptors to prevent this for specific cases, it is hard to imagine a solution to this problem in its entirety without having specific categories or descriptors for every single robot, defeating the purpose of a more generalized model to describe appearance.}

\replaced[id=rn]{Many studies \cite{ABOT,Löffler2020a,kalegina2018characterizing,seifi2023firsthand} compiled datasets of robots and their features to examine patterns in how people perceive and ascribe characteristics to robots based on their appearance.}{
\textcite{ABOT}, \textcite{kalegina2018characterizing}, and \textcite{seifi2023firsthand} used datasets that contained lists of robot features compiled during their research, analyzing how various traits are attributed to different types of robots. By correlating these traits with specific visual features in their datasets, they could examine patterns in how people perceive and ascribe characteristics to robots based on their appearance.} The \textsc{MetaMorph} model, along with the comprehensive dataset of robot appearance features that has been compiled, will support similar studies on a larger scale as it includes a broader range of detailed features and descriptors. A follow-up study of this work exploring how participants perceive similarity between the different subdivisions and descriptors would enable more accurate weighted visual distance calculation between robots.

\begin{wrapfigure}[12]{r}{0.2\textwidth}
  \vspace{-1.5em}
  \begin{center}
    \includegraphics[width=0.2\textwidth]{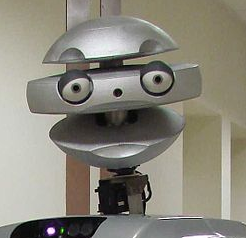}
  \end{center}
    \vspace{-0.5em}
  \caption{Portrait of \emph{Flash} \cite{kedzierski182budowa}; adapted from \cite{FlashImage}.}
  \label{figflash}
\end{wrapfigure}

During the validation, it became clear that it is problematic to describe morphological features that emerge symbolically from the interaction of different morphological features using the \textsc{MetaMorph} model. Examples are the EMYS \cite{EMYS} and Flash \cite{kedzierski182budowa} robots developed by the Wroclaw University of Technology\replaced[id=rn]{, which both}{. Both models} share a head divided into three segments where facial features like the mouth and eyebrows are represented by the space between segments, as seen in \autoref{figflash}. A solution would require a strict separation of physical form from their morphological interpretation by humans. However, this would complicate the model to the extent that we fear that adoption might be limited. Since our model is based on interpretation, e.g., two cameras placed in the upper face section would be considered symbolic eyes since they were interpreted as such by the coders, it would be questionable if this additional purely physical description would actually lead to any benefits and essential additional information.

Participants in the focus group mentioned that they approved of the model describing the spatial relations between the different components, but considered the split into the four different zones too simplified. While we considered various ways of describing the location of components on the robot during the development of the model, we elected to omit them from the final model since the established URDF format already contains information on the exact spatial location of the different robot components. It could be beneficial to suggest an extension of the format where, in addition to the \texttt{<visual>}-tag that describes a joint by providing a 3D model, a complimentary tag could contain the concepts applying to this joint according to \textsc{MetaMorph}.

\section{Limitations and Future Work}
Our work presents important considerations for the classification of robot appearances through the development of \textsc{Meta-Morph}. Still, certain limitations should be acknowledged.

First, our dataset is derived exclusively from the IEEE Robots Guide, which may limit the diversity of robot appearances included in the model. In addition, we elected to accept their definition of what is considered a robot. However, there were some entries where this classification was questionable, e.g., the Watson computer system or the Replicator+ 3D printer. Future research should consider expanding the dataset by incorporating robots from other sources.
This would allow for a more comprehensive validation of \textsc{MetaMorph} across a broader range of robot designs and possibly enrich the model with additional descriptors if needed.
Importantly, the version of the model presented in this paper is not final. While this is a first step towards a comprehensive model of robot appearance, the options given in the taxonomy only reflect features present in the robots that are part of the IEEE Robots Guide. Our work aims to provide guidance on how to systematically describe robot appearance while allowing for and encouraging the extension of options such as the number of joints or the variety of tools in the future as necessary.

In our case, step 3 of the metamodeling process (the concept and relation extraction) and step 7 (the metamodel validation) were performed by two researchers, with an additional independent expert deciding on conflict cases. Since our model is based on people's perception and interpretation of the features, having this step repeated by a larger group of people - maybe even laypeople instead of experts on HRI or robotics - might be needed to further validate the coding results produced by the experts as being representative of the general public.

It should also be acknowledged that the \textsc{MetaMorph} model only describes robot appearance and does not consider robot capabilities. While we do feel that capability and function are important factors that, at least in some cases, dictate the appearance and should be considered when assessing the interaction, cases could arise where the appearance does not necessarily match the function, e.g., a robot appendage might be shaped like a hand and influence the interaction by being perceived as one, while being unable to actually realize functions like grasping. In line with existing work, such as the taxonomy to describe HRI by \textcite{onnasch2021taxonomy}, we suggest using the \textsc{MetaMorph} model to detail robot appearance and describe other factors such as robot capability, movement, speech, or behavior separately. A similar metamodeling approach could be applied to these areas to extend the \textsc{MetaMorph} model with additional dimensions and provide a more detailed description of these factors.

\added[id=rn]{On the same note, other factors that affect not only HRI in general, but potentially also how appearance shapes HRI, are not considered by the model, e.g., culture, acceptance of robots, and ethics. Studies have to consider these separately.}

Finally, creating a website to present the \textsc{MetaMorph} model and dataset would enhance its visibility and accessibility. The website could offer search and filtering options, enabling users to explore the dataset easily. Moreover, an application that allows users to create and customize robot models based on the described features and export their designs would provide a practical tool for researchers and practitioners. We plan to address these points in the near future, aiming to enhance the robustness and accessibility of \textsc{MetaMorph}.




\section{Conclusion}
In this work, we evaluated existing frameworks for classifying robot appearances and identified a significant gap: the absence of a comprehensive framework encompassing a wide range of robot designs. In response, we developed the \textsc{Meta\-Morph} model via a metamodeling approach, which extends beyond anthropomorphic and zoomorphic robots and provides a structured method for classifying and comparing visual features across all robot types. Our contribution facilitates systematic comparisons across various experimental conditions, improving the clarity and consistency of studies and literature reviews. The \textsc{MetaMorph} model also provides data as a basis for calculating the visual distance between robot models, enabling quantitative measurement for assessing the similarities and differences in their appearances in the future. Additionally, it serves as a foundation for exploring desirable visual traits in robots, helping researchers and designers tailor robot appearances to specific tasks and contexts.



\section*{Acknowledgment}
 We thank Iddo Wald, the focus group participants, and the anonymous reviewers for their feedback, which has greatly contributed to improving this work.


\printbibliography
\end{document}